\documentclass[10pt,conference]{IEEEtran}
\usepackage[style=ieee]{biblatex} 
\usepackage{pifont}
\usepackage{graphicx}      % Include graphics
\usepackage{amsmath}       % Math environments
\usepackage{amssymb}       % More math symbols
\usepackage{algorithmic}   % Algorithms
\usepackage{algorithm}     % Algorithm float
\usepackage{textcomp}      % Extra symbols
\usepackage[T1]{fontenc}          % T1 encoding – required for Times
\usepackage{newtxtext}            % Times text
\usepackage{newtxmath}            % Times math (bold math, etc.)

\usepackage{changepage}
\definecolor{ieeeblue}{HTML}{00629B}
\definecolor{myred}{RGB}{139,0,0}   % darker red
\usepackage[breaklinks,colorlinks,allcolors=ieeeblue]{hyperref}
\usepackage[normalem]{ulem}
\usepackage{tabularray}
\usepackage{siunitx}
\UseTblrLibrary{siunitx}

\usepackage[capitalise]{cleveref}

\usepackage[format=plain,labelformat=simple,labelsep=period,font=small,compatibility=false]{caption}
\usepackage[font=footnotesize,skip=3pt,subrefformat=parens]{subcaption}

\usepackage{silence}  % Suppress unwanted warnings
\title{Getting the Numbers Right---Modelling Multi-Class Object Counting in Dense and Varied Scenes}

\author{
    \IEEEauthorblockN{
        Villanelle O'Reilly\IEEEauthorrefmark{1}\IEEEauthorrefmark{4},
        Jonathan Cox\IEEEauthorrefmark{1},
        Georgios Leontidis\IEEEauthorrefmark{2},
        Marc Hanheide\IEEEauthorrefmark{1},
        Petra Bosilj\IEEEauthorrefmark{3},
        James M. Brown\IEEEauthorrefmark{4}
    }
        
    \IEEEauthorblockA{\IEEEauthorrefmark{2}Interdisciplinary Institute, University of Aberdeen, UK \& UiT The Arctic University of Norway}
    
    \IEEEauthorblockA{\IEEEauthorrefmark{3}Department of Advanced Computing Sciences, Maastricht University, The Netherlands}

    \IEEEauthorblockA{\IEEEauthorrefmark{1}L-CAS, University of Lincoln, UK; \IEEEauthorrefmark{4}AVAIL, University of Lincoln, UK}

    \IEEEauthorblockA{\ding{41} {\tt\small \{voreilly, jcox, mhanheide, jamesbrown\}@lincoln.ac.uk} \ding{41} {\tt\small  georgios.leontidis@uit.no} \ding{41} \\
        \ding{41} {\tt\small petra.bosilj@maastrichtuniversity.nl} \ding{41}
    }
}

\begin{document}
    \maketitle

    %---------------------------------------------------------------
    % Abstract
    %---------------------------------------------------------------
    \begin{abstract}
        Density map estimation enables accurate object counting in heavily occluded, and densely packed scenes where detection-based counting fails. In multi-class density estimation, class awareness can be introduced by modelling classes non-exclusively, better reflecting crowded and visually ambiguous contexts. However, existing multi-class density estimators often degrade in less-dense scenes, while state-of-the-art detectors still struggle in the most congested settings. To bridge this gap, we propose the first vision-transformer-based approach to multi-class density estimation. Our model combines a Twins-SVT pyramid vision transformer backbone with a multiscale CNN decoder that leverages hierarchical features for robust counting across a wide range of densities. Further to that, the method adds an auxiliary segmentation task with the \emph{Category Focus Module} to suppress inter-category interference at training time. The module improves the density estimation head without the need for constraining assumptions added by the application of the auxiliary task at inference time, as required in previous methods. Training and evaluation on the VisDrone and iSAID benchmarks demonstrates a leap in performance versus the previous state-of-the-art multi-class density estimation methods, attaining a 33\%, 43\%, and 64\% reduction to MAE in testing evaluation. The method outperforms YOLO11 in less busy scenes, exceeding it by an order of magnitude in the most crowded testing samples.
        
        Code, and trained weights available at \href{https://github.com/LCAS/gnr_mcdest}{gh:LCAS/gnr\_mcdest}.
    \end{abstract}

    %---------------------------------------------------------------
    % Keywords
    %---------------------------------------------------------------

    %---------------------------------------------------------------
    % Main sections
    %---------------------------------------------------------------
    \section{Introduction}
\label{sec:intro}

A fundamental computer vision task, object counting produces valuable automated insights for urban planning \cite{shen2019pedestrian,liang2022trans}, biodiversity monitoring \cite{hicks2021deep, ocer2020tree} and in medicine \cite{lavitt2021blood}. It is an expensive and tedious task for humans to perform at scale \cite{breeze2021pollinator}, and the high density of objects in occluded scenes pose a significant challenge for automated methods. Computer vision object counting methods include counting by detection \cite{hicks2021deep}, direct regression of object counts, without localisation information \cite{liang2022trans}, and density estimation providing ``weak'' localisation, directly regressing a heatmap that integrates into the total counts of objects \cite{liu2021context, dong2024weakly}.

\begin{figure}
    \centering
    \includegraphics[width=0.7\columnwidth]{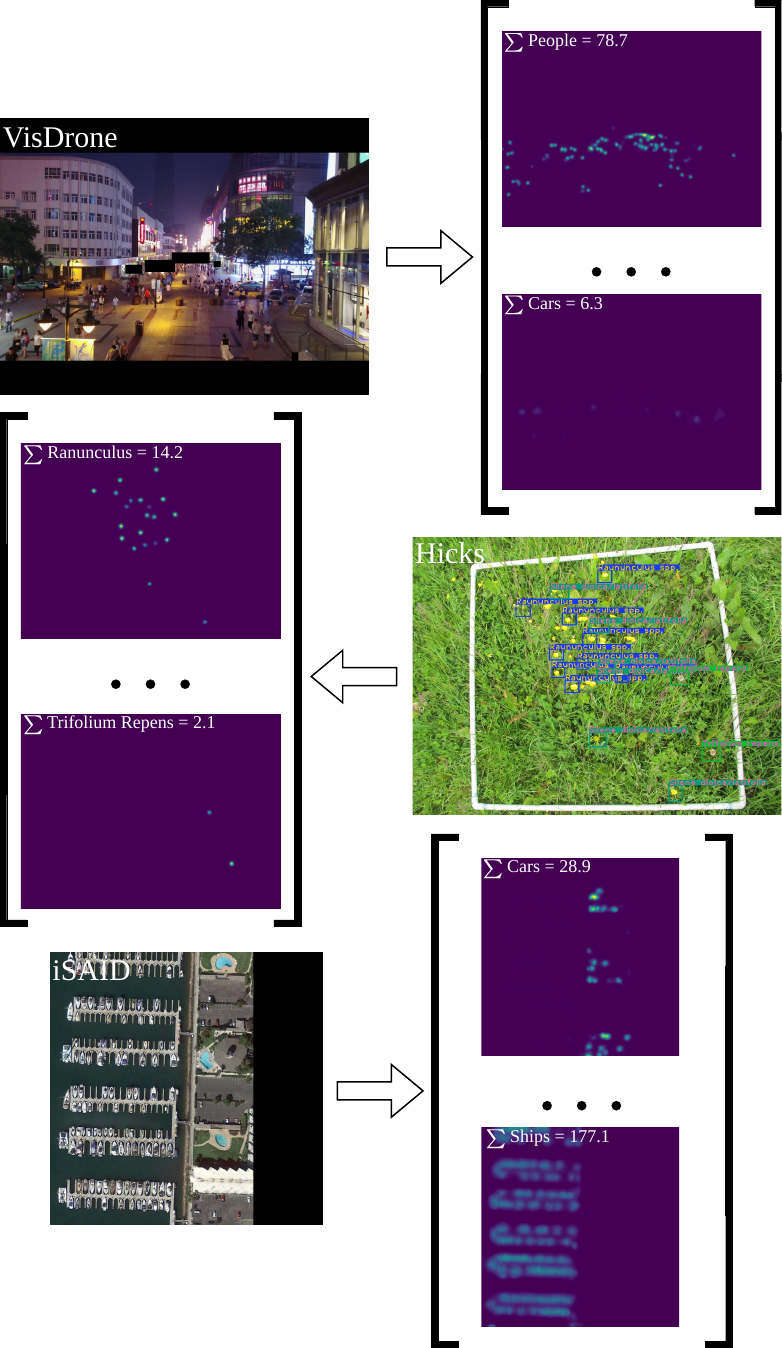}
    \caption{\textbf{Multi-class Density Estimation}. The proposed model's inference on testing samples from the \textcite{hicks2021deep}, VisDrone-DET \cite{zhu2021detectionVisDrone}, and iSAID \cite{waqas2019iSAID} datasets. The estimated count, from the integration of the heatmap \cref{eq:count_est}, is superimposed category-wise.}
    \label{fig:multipurpose}
    \vspace{-6mm}
\end{figure}

Density estimation can be extended to a multi-class paradigm, where one density map is predicted for a fixed set of object classes, and is rarely studied \cite{xu2021dilated,michel2022class,fu2024dense,gao2024nwpu}. While multi-class object detection has seen extensive research \cite{zou2023detection}, the continuous density estimation model yields more accurate counts in crowded and occluded scenes than detection methods, as demonstrated in \textcite{gomez2021deep}. \emph{Crowd counting} methods, first applied to estimating the population of congested urban streets \cite{liu2021context}, can be directly applied for counting singular abstract classes, such as agricultural items (i.e. one of: almonds, apples, wheat kernels, etc.) in \textcite{gomez2021deep}. More recent developments see density estimation used in an open-vocabulary textual prompting paradigm, but the state-of-the-art has been found to lack class-awareness in \textcite{ciampi2025mindtheprompt}, and such methods lack the single-stage design of multi-class density estimation---as only one class or prompt can be isolated at a time.

\begin{figure}
    \centering
    \includegraphics[width=1\columnwidth]{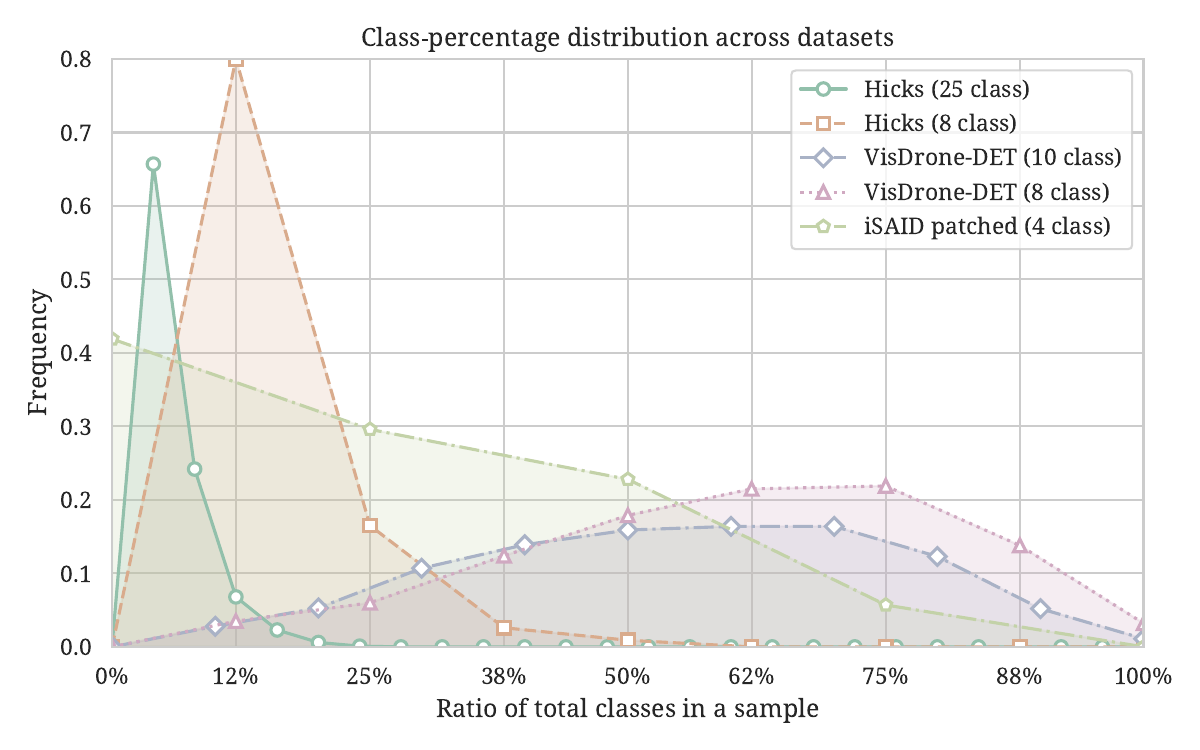}
    \caption{\textbf{Class-Appearance Distribution}. The plot illustrates, across several datasets, the distribution of the proportion of classes present within a single sample. VisDrone-DET and iSAID images typically contain many classes, whereas biodiversity samples from Hicks generally contain no more than one class in a sample. \vspace{-0.5cm}}
    \label{fig:intro:classdist}
\end{figure}
Multi-class methods simultaneously count several categories of objects in one pass, such as people, cars, trucks in DSACA \cite{xu2021dilated} or additionally cars, trucks, ships and planes from satellite imagery in \textcite{michel2022class}. Previous multi-class density estimation methods \cite{xu2021dilated,michel2022class,fu2024dense,gao2024nwpu} have only been evaluated on urban datasets including VisDrone-DET \cite{zhu2021detectionVisDrone}, iSAID \cite{waqas2019iSAID} and RSOC \cite{gao2021countingRSOC}, datasets containing either few classes, or a uniform class-frequency distribution. Visualised in \cref{fig:intro:classdist}, the biodiversity dataset \textcite{hicks2021deep} is extremely skewed, compared to the more uniformly distributed VisDrone \cite{zhu2021detectionVisDrone} and iSAID \cite{waqas2019iSAID} datasets. It is common for most of the categories to appear in a single sample in the urban datasets, contrasting the rare occurrence of more than one category present in \textcite{hicks2021deep}, where some species of flowers may never naturally occur next to another. We find our vision-transformer-based method is able to successfully learn in this new domain, beating the previous open-source state-of-the-art DSACA \cite{xu2021dilated}, which we train with the dataset.

Contributions: \textbf{1)} The first vision-transformer-based multi-class density estimation architecture, outperforming the state of the art with an average $46.7\%$ MAE decrease over the two benchmarks, and outperforming the object detection YOLO11 method by an order of magnitude in the most occupied scenes. We evidence the robustness of the method with the superior counting predictions over YOLO11 in less busy scenes that are historically favourable to object detection. \textbf{2)} Fusing a Convolutional Neural Network (CNN) decoding features from the Vision Transformer (ViT) Twins-SVT \cite{chu2021twins} backbone, the novel multiscale multi-class counting head is built around the smoothly-differentiable and numerically stable softplus activation, guaranteeing directly explainable density estimations modelling positive counts. \textbf{3)} The Category Focus Module, used in a lightweight auxiliary masking task, further increases class awareness, and directly leverages the density output at inference time. To the best of our knowledge, this is the first multi-class object counting approach to cast the task entirely as continuous density estimation at inference time, avoiding the discreteness assumptions inherent to detection-augmented methods \cite{michel2022class}, and the effectively thresholding object-presence-likelihood assumptions implicit in segmentation-augmented approaches \cite{xu2021dilated}.

    \section{Related Work}
\label{sec:related}
\emph{\textbf{YOLO}}: YOLO11 is a recent YOLO architecture \cite{yolo11_ultralytics,khanam2024yolov11}, and the state-of-the-art in single-stage object detection. We use it as a benchmark for counting-by-detection performance.

\emph{\textbf{DSACA}}: DSACA \cite{xu2021dilated} was the first method reformulating crowd counting into a multi-class counting and density estimation problem. It extracts multiscale features from a VGG-16 \cite{simonyan2015vgg} backbone with a Dilated Scale Aware Module (DSAM). They suppress inter-category interference with their Category Attention Module (CAM) masking out \emph{low confidence} regions of the output density map at inference time. DSACA achieved state-of-the-art performance on the multi-class VisDrone-DET \cite{zhu2021detectionVisDrone}, and RSOC \cite{gao2021countingRSOC} datasets, benchmarking against various \cite{zhang2016MCNN,cao2018scale,li2018CSRNet,liu2021context,ma2019bayesian} single-class density estimation methods.

% The multi-task approach allows the gradients from the CAM backpropagation to improve the "classification" capabilities of the DSAM counting branch as the backbone weights remain live.\jamessnote{not a very clear sentence}

\begin{figure*}[t]
    \centering
    \includegraphics[width=0.70\linewidth, trim=0 0 24mm 0, clip]{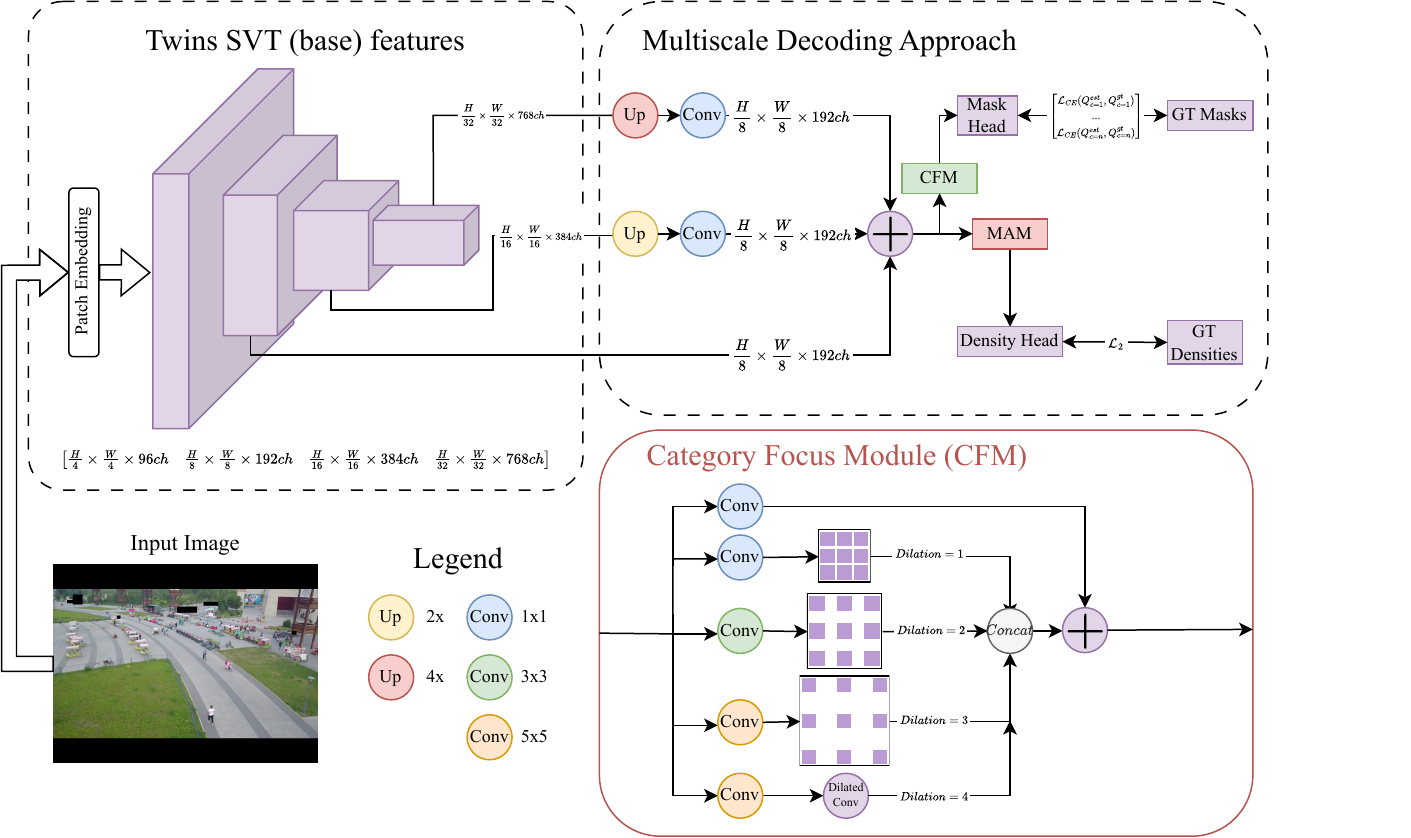}
    \caption{\textbf{The Model Architecture}. In the Category Focus Module (CFM), an extension of the Multiscale Aware Module (MAM) from MRCNet \cite{yu2025multiscale}, the first column of convolutions is followed by a column of batch norm and ReLU activations. The same batch norm and ReLU column is present in the MAM.}\label{fig:model:main}
    \vspace{-4mm}
\end{figure*}

\emph{\textbf{Class-aware Object Counting}}: \textcite{michel2022class} employ a CNN multitask architecture with a ResNet50 \cite{he2016resnet} backbone. A region proposal network, and a detection head generate class-aware detections, alongside a parallel multi-class density-estimation branch that learns a separate density map for each category. Therefore, the key difference from DSACA \cite{xu2021dilated} is the use of an object-detection-based auxiliary task rather than a segmentation-based one. The two branches are fused at inference time in a Count-Estimation Network (CEN), which predicts integer object counts by abstracting the continuous density maps, and suppressing inter-category interference by combining density-map and detection information. This design offers another way to reduce inter-category interference via multitask learning, combining a task suited to low-density scenes with a task suited to high-density scenes, and achieving reasonable counting performance across both count ranges. \textcite{michel2022class} evaluate their method against object-detection approaches \cite{ren2017fasterrcnn,qiao2021detectors,duan2019centrenet,he2017maskrcnn} and report state-of-the-art results on those datasets.

\emph{\textbf{CCTwins}}: \textcite{dong2024weakly} propose a weakly-supervised ViT and CNN based single-class crowd counting method that relies solely on count-level labels rather than dense location annotations. Building on the Twins-SVT backbone \cite{chu2021twins}, it introduces an adaptive scene-consistency attention module to enhance feature extraction in highly uneven crowd distributions, and a multi-level weakly-supervised loss that progressively refines the density map from coarse to fine stages. The method uses a multiscale convolutional decoder repeatedly fusing multiscale feature representations, supervising each layer against the global count. Rooted in their success concatenating and convolving multiple layers of the pyramid-ViT backbone, we take the Twins-SVT backbone for our multi-class method.

\emph{\textbf{MRCNet}}: As a recent publication with state-of-the-art fully-supervised performance, we draw from \textcite{yu2025multiscale} as the most modern method for a density estimation multiscale feature extraction decoder. As a modern technique not using vision transformers, they introduce a dilated \emph{Multiscale Aware Module} increasing the receptive field of their decoder. To increase our architecture's robustness to scale and density variance, we adopt this component of MRCNet.

For our density estimation head, we use a similar multiscale convolutional decoding scheme, leveraging hierarchical features as with CCTwins \cite{dong2024weakly} and MRCNet \cite{yu2025multiscale}, for scale-variance-robust feature extraction capabilities from the pyramid ViT Twins-SVT backbone; introducing self-attention to multi-class density estimation. Similar to DSACA \cite{xu2021dilated}, we employ a segmentation-based secondary task to reduce inter-category interference. We extend their two-task design by introducing the \emph{Category Focus Module} that extracts features from several layers of the backbone, learning at multiple scales with its \emph{Multiscale Aware Module-style} design. By designing our lightweight masking head, we rely on the unified propagation of the segmentation and density losses as the sole benefit of the masking task, realised at training time, without the need for the masking head at inference. By not applying the assumptions and issues a segmentation \cite{xu2021dilated} or object detection \cite{michel2022class} task has in dense and occluded scenes, where the tasks could lead to exclusion of correct density contributions, ours is the first multi-class density estimation method that truly operates with a continuous model designed for such contexts.
    \section{Method}
\label{sec:method}
Following previous work \cite{michel2022class,xu2021dilated}, we formulate multi-class density estimation as learning to predict a set of density maps per input image, where each map represents the spatial distribution of a single class in the image. Let $D^\text{est}_i$ denote the predicted set of density maps estimated from the RGB input image $I_i \in \mathbb{R}^{H \times W \times 3}$. We assume a non-linear mapping $\mathcal{F}$, parameterised by $\theta$, such that:
\begin{equation}
    D_{i}^{\text{est}}(I_i) = \mathcal{F}(I_i; \theta)
    \label{eq:denest}
\end{equation}
For a dataset of $N$ images $I = \{ I_1,\; I_2,\; \dots,\; I_N \}$, $\theta$ is learned using $N$ ground-truth density maps, each containing $C$ channels representing the set of classes in a dataset. The estimated and ground-truth maps share the same spatial dimensions, discretised at $\frac{1}{4^2}^{\text{th}}$ the resolution of $I_i$. Thus, the maps have the width $W_d = \frac{W}{4}$, and the height $H_d = \frac{H}{4}$, yielding $D^{\text{gt}}_i, D^{\text{est}}_i \in \mathbb{R}^{H_d \times W_d \times C}$.

We index the classes of $D^{\text{est}}_i$ as $\{ D^{\text{est}}_{i,1}, \dots, D^{\text{est}}_{i,C} \}$. For example, $D^{\text{est}}_{i,1}$ might represent the \emph{people} class for VisDrone, or the \emph{planes} class for iSAID.

To determine the count of objects for the class $c$ present in the input image $I_i$, the density estimation for all classes is predicted from the function $\mathcal{F}$ \cref{eq:denest}, from which the corresponding channel $c$ in the density prediction is isolated as $D^\text{est}_{i,c}$. The estimated count of objects for the class $c$, defined as $\hat{n}_{i,c}$, is the integration of $D^\text{est}_{i,c}$, and the ground-truth count, $n_{i,c}$, is the integration of $D^\text{gt}_{i,c}$ such that:
\begin{subequations}\label{eq:counts}
    \allowdisplaybreaks[1]
    \begin{align}
        \hat{n}_{i,c}
        &= \sum_{w=1}^{W_d}\sum_{h=1}^{H_d}
            D^{\text{est}}_{i,c,w,h},                      \label{eq:count_est} \\[4pt]
        n_{i,c}
        &= \sum_{w=1}^{W_d}\sum_{h=1}^{H_d}
            D^{\text{gt}}_{i,c,w,h}.                       \label{eq:count_gt}
    \end{align}
\end{subequations}
Our method implements $\mathcal{F}$ using a Pyramid Vision Transformer (PVT) backbone coupled with a Convolutional Neural Network (CNN) decoder (\cref{sec:method:model}). The parameters $\theta$ are optimised iteratively using the loss functions defined in \cref{sec:method:loss}. % The CNN parameters are initialised by a pseudorandom process, and parameters belonging to the Twins-SVT \cite{chu2021twins} PVT backbone are initialised directly from a Twins ``ImageNet-1k'' classification model.

\subsection{Backbone and Decoding Approach}\label{sec:method:model}
    Many density estimation methods \cite{yu2025multiscale,lin2025distribution,xu2021dilated,fu2024dense}, use the CNN VGG16 \cite{simonyan2015vgg} backbone, or the ResNet50 \cite{michel2022class} backbone. A number of density and count-estimation methods using vision transformer \cite{lin2025distribution,liang2022trans}, or pyramid vision transformer backbones \cite{tian2021cctrans,dong2024weakly} have emerged with greater single-class performance than methods using a CNN backbone. Therefore, we use the Twins-SVT \cite{chu2021twins} backbone, combined with a multiscale CNN-based decoder. By using a pyramid vision transformer with spatial attention, as opposed to a flat transformer encoder (e.g. \cite{liang2022trans}), we are still able to extract multiscale features enabling the convolutional decoder to learn scale variances expected in problems suited to density estimation.

\begin{figure}
        \vspace{-.15cm}
        \centering
        \includegraphics[width=0.9\linewidth, trim=0 0 7mm 0, clip]{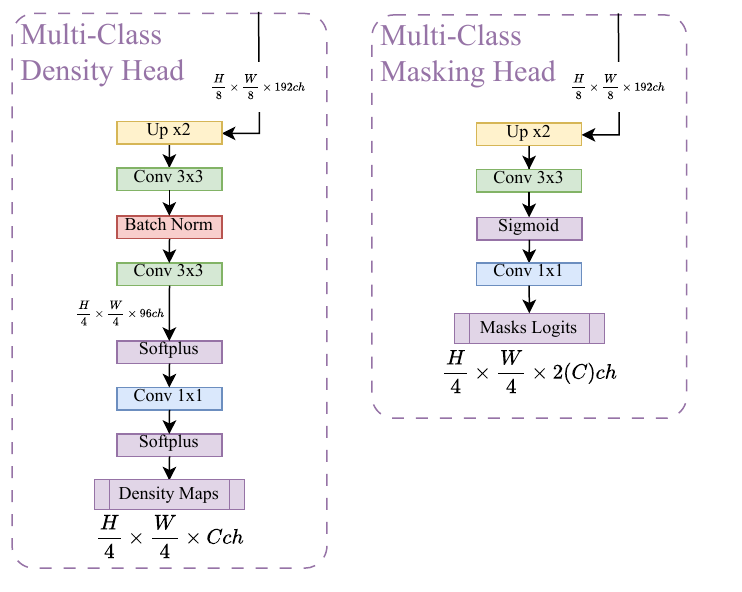}
        \caption{\textbf{Model Heads}. The two output heads of the model.}\label{fig:model:heads}
        \vspace{-.6cm}
    \end{figure}

    As we are interested in the mask's cross-entropy loss improving performance of the density head, we present a minimal masking head design in \cref{fig:model:heads}, so that the cross-entropy segmentation loss has a greater impact on the density head. Our \emph{Category Focus Module} (CFM), combines the approaches of DSACA \cite{xu2021dilated} and CCTrans \cite{tian2021cctrans}, passing fused features from layers of the encoder through dilated convolutions to improve robustness to scale variation by increasing the receptive field. The masking head outputs $2C$ channels of logits, so that for each class there is a segmentation task not mutually exclusive of other classes, predicting a positive and background mask for each class. This corresponds to the assumption that any given pixel could be representative of more than one class. The \emph{Multiscale Aware Module} similarly processes features from the backbone at different scales with dilated convolutions to increase the receptive field of the density head, to the end of scale-robust density predictions.
    
    \cref{fig:model:heads} shows the class-aware density head enabling the method to simultaneously predict several classes of density maps. The approach uses the smoothly differentiable softplus activation:
    \begin{equation}
        \text{softplus}(x) = \frac{\log(1 + \exp(\beta \times x))}{\beta}.
        \label{eq:sp}
    \end{equation}
    Softplus provides a bounded output, as with ReLU (used in \cite{xu2021dilated,michel2022class} etc.), constraining the predictions of the model to positive numbers. However, where ReLU would truncate negative predictions before the calculation of a loss, the softplus function guarantees a numerically stable positive output, applying our counting constraint without sacrificing gradients at training time.

\subsection{Loss Functions} \label{sec:method:loss}
    We use the $\mathcal{L}_2$ loss for optimising the density estimation task. This computes the class-pixel-wise squared error, therefore incorporating spatial and class-localisation error, as well as count error into the optimisation problem:
    \begin{equation}
        \mathcal{L}_2 \big( D^{\text{gt}}_{i}, D^{\text{est}}_{i} \big) = \sum^{C}_{c=1} \sum^{H_d}_{h=1} \sum^{W_d}_{w=1} \big( D^{\text{gt}}_{i,c,w,h} - D^{\text{est}}_{i,c,w,h} \big)^2.
        \label{eq:l2}
    \end{equation}
    For our auxiliary function, the masking task employs a per-category cross-entropy loss, where $Q^\text{gt}$ and $Q^\text{est}$ refer to the ground-truth and estimated mask data:
    \begin{equation}
        \mathcal{L}_{mask} = \frac{1}{C} \sum^C_{c=1} \mathcal{L}_{CE}(Q^\text{est}_{c}, Q^\text{gt}_{c}).
        \label{eq:lmask}
    \end{equation}
    The losses are combined with the weighting factor $\omega$. This enables the parameters $\theta$, representing the lightweight masking task, the density head, and the parameters shared between the tasks to be optimised in one back-propagation operation, creating a dynamic relationship between the tasks in the shared parameters:
    \begin{equation}
        \mathcal{L} = (\omega \times \mathcal{L}_{mask}) + \mathcal{L}_\text{2}.
        \label{eq:lall2}
    \end{equation}

\subsection{Ground-truth Generation}
    As our datasets provide bounding box annotations, we take centroid points of the bounding boxes and apply an empirically-derived series of Gaussian kernels to generate smooth density maps, via the same procedure as DSACA \cite{xu2021dilated}. As the radius of the kernel can exceed the bounds of the image, which would fractionally subtract from the represented object count, the density maps are scaled class-wise to ensure that the integral of the density map produces the same integer count of objects as the bounding box data provides.
    \section{Experiments}
\label{sec:experiments}

% --- Macro for column type with centering on the decimal place ---
\NewColumnType{C}{X[c,m,si={table-format=2.2, table-space-text-post={-}, detect-weight=true, mode=match}]}

\begin{table*}[t!]
  \centering
  \begin{tblr}{width=\textwidth,
      colspec={Q[c,m] Q[c,m] | C C | C C | C C C | C C},
      hline{1,Z} = {2pt},
      hline{2,3} = {1pt},
      hlines}
    \SetCell[c=11]{c} VisDrone-DET (8-category) \\
    \SetCell[c=2]{c,m} {\textbf{Count Range} \\ (samples in range)} & & \textbf{YOLO 11l \cite{yolo11_ultralytics}} & {{{\textbf{YOLO 11x \cite{yolo11_ultralytics}}}}} & {{{\textbf{Ours (no CFM)}}}} & {{{\textbf{Ours (single-scale CFM)}}}}   & {{{\textbf{Ours (CFM, Twins-SVT-small)}}}} & {{{\textbf{Ours (CFM, Twins-SVT-base)}}}} & {{{\textbf{Ours (CFM, Twins-SVT-large)}}}} & {{{\textbf{Ours (CFM, ReLU)}}}} & {{{\textbf{Ours (CFM, leaky ReLU)}}}} \\

    \SetCell[c=2]{c}\textbf{No. Params}
      &  & 25.37m & 56.97m &  58.20m & 60.95m &  26.23m & 60.95m & 107.95m & 60.95m & 60.95m \\

    \SetCell[r=2]{c}{\textbf{0--1000}\\(1610)}
      & \textbf{MAE}   & 2.75 & 2.65 & 2.49 & \uline{2.28} & 2.38 & 2.29 & \textbf{2.27} &  5.83 &  5.67 \\
      \SetHline[1]{2-11}{dashed}
      & \textbf{RMSE}  & 10.49 & 10.06 & \textbf{8.00} & 8.16 & \uline{8.06} & 8.28 & 8.18  & 16.30 & 10.38 \\
      
    \SetCell[r=2]{c}{\textbf{0--10}\\(126)}
      & \textbf{MAE}   & 1.85 & 1.51 &  0.63 & 0.53 & 0.59 & \uline{0.46} & \textbf{0.42} &  0.85 &  3.97 \\
      \SetHline[1]{2-11}{dashed}
      & \textbf{RMSE}  & 2.67 & 2.44 & 1.04  & 0.99 & \uline{0.98} & 1.08 & \textbf{0.83}  &  1.99 &  4.36 \\

    \SetCell[r=2]{c}{\textbf{11--50}\\(980)}
      & \textbf{MAE}   & 7.81 & 7.49 & 1.62  & 1.41 & 1.51 & \uline{1.39} & \textbf{1.37} &  3.62 &  4.79 \\
      \SetHline[1]{2-11}{dashed}
      & \textbf{RMSE}  & 10.83 & 10.46 & 3.10 & \textbf{2.78} & 2.80 & \uline{2.79} & 2.81  &  7.35 &  5.86 \\

    \SetCell[r=2]{c}{\textbf{51--100}\\(377)}
      & \textbf{MAE}   & 23.47 & 22.21 & 3.31 & \textbf{3.06} &  3.12 & 3.09 & \uline{3.09} &  8.40 &  6.39 \\
      \SetHline[1]{2-11}{dashed}
      & \textbf{RMSE}  & 28.35 & 27.29 & 6.59 & \uline{6.32} &  \textbf{6.22} & 6.49 & 6.49  & 16.46 &  9.32 \\

    \SetCell[r=2]{c}{\textbf{101--1000}\\(127)}
      & \textbf{MAE}   & 84.94 & 80.63 & \uline{8.55} & \textbf{8.49} &  8.63 & 8.61 & 8.61 & 20.23 & 12.06 \\
      \SetHline[1]{2-11}{dashed}
      & \textbf{RMSE}  & 109.41 & 105.17 & \textbf{24.64}  & 25.78 &  \uline{25.45} & 26.12 & 25.73 & 46.31 & 28.69
  \end{tblr}
  \caption{Test-set counting results on the merged 8-category VisDrone-DET \cite{zhu2021detectionVisDrone} dataset, ablating the backbone, Category Focus Module (CFM), and softplus activation. Best and second-best results per row are in \textbf{bold} and \uline{underlined}, respectively (ranked using unrounded metrics). For comparison, the VGG-16 backbone used in DSACA \cite{xu2021dilated}, and by \textcite{michel2022class} has 138m parameters, making those baselines larger than any of our variations. Unless noted otherwise, models use the Twins-SVT-base backbone (60.95m parameters) and softplus activation. Metrics are grouped by total per-image object counts: '0--1000' represents the full VisDrone-DET test set, while '11--50' includes samples where the total ground-truth count $n_{i,c}$ is $11 \leq \sum^C_{c=1} n_{i,c} \leq 50$.}
  \label{table:ablations}
\end{table*}

We assess our method's performance with the evaluation on the benchmark datasets VisDrone \cite{zhu2021detectionVisDrone} and iSAID \cite{xia2018DOTA,waqas2019iSAID}. We further apply our method to the biodiversity monitoring problem with data from \textcite{hicks2021deep}. A sample for each of the datasets is visualised in \cref{fig:multipurpose}, along with some classes of our model's predicted density maps.

\subsection{Datasets}

    \emph{\textbf{VisDrone-DET \cite{zhu2021detectionVisDrone}}}: Collected over 14 different Chinese cities, the VisDrone \emph{Object Detection in Images} dataset contains 10,209 images of varying density, with 10 annotated classes including 8 types of vehicle, and people. \textcite{michel2022class} evaluate against the full 10-class VisDrone-DET dataset. As DSACA \cite{xu2021dilated} merge the \emph{people} and \emph{pedestrian} classes into a unified people class, and the \emph{tricycle} and \emph{awning-tricycle} classes into a unified tricycle class, their VisDrone-DET sub-dataset has 8, instead of the full 10 classes used in \textcite{michel2022class}. Therefore, we evaluate our model against both the 8-class and 10-class VisDrone-DET datasets. Before dataset ground-truth density generation, we resize the images to be 1024 pixels wide, proportionally scaling the height.

    \emph{\textbf{iSAID (4-class) \cite{waqas2019iSAID}}}: As described in \textcite{michel2022class}, patch the iSAID dataset into $800 \times 800$ pixel tiles. In the absence of a 4-class or crowd-counting-compatible test challenge, we combine, shuffle, and split the public validation and training datasets 70--10-20 to training-validation-testing. We split the dataset before patching, so that no single image can be patched across the subsets, which could inflate the testing results. Without a provided 4-class iSAID test set, there will be a margin for error comparing our method to \textcite{michel2022class}, although it's expected to be minimal because of the large size of the dataset at 20,906 patches, and as we copy their dataset preparation procedure. The YOLO11 model is trained and evaluated on the same split of the dataset we produce for density estimation, before the generation of the ground-truth density maps.

    \emph{\textbf{\textcite{hicks2021deep} (8-class)}}: A biodiversity monitoring dataset, the Hicks dataset of annotates 25,352 tags between 25 species of flowers in natural environments. We reduce the biodiversity monitoring problem to the 8 most common species of flowers. The images in the dataset are resized to have a maximum width of 1024 pixels, and where scaled to reduce the width the height is scaled proportionally. The validation and training sets are combined and split 70--10-20 to training-validation-testing, as the 50-image testing set of the dataset does not contain annotations.

\begin{table}[b!]
  \centering
  \begin{tblr}{width=\linewidth,
              colspec={Q[c,m] Q[c,m]|C C},
              hline{1,Z} = {2pt},
              hline{2,3} = {1pt},
              hlines}
    \SetCell[c=4]{c} \SetCell{c} VisDrone-DET (10-category) \\
    \SetCell[c=2]{c} {\textbf{Count Range} \\ (samples in range)} & & \SetCell{c, font=\bfseries} {{{\citeauthor{michel2022class} \cite{michel2022class}}}} & \SetCell{font=\bfseries} {{{Ours}}} \\

    \SetCell[r=2]{c}{\textbf{0--1000}\\(1610)}
      & \textbf{MAE}  & 3.76 & \SetCell{font=\bfseries} 1.99 \\
      \SetHline[1]{2-4}{dashed}
      & \textbf{RMSE} & 9.56 & \SetCell{font=\bfseries} 6.41 \\
      
    \SetCell[r=2]{c}{\textbf{0--10}\\(126)}
      & \textbf{MAE}  & 2.07 & \SetCell{font=\bfseries} 0.52 \\
      \SetHline[1]{2-4}{dashed}
      & \textbf{RMSE} & 3.25 & \SetCell{font=\bfseries} 0.96 \\

    \SetCell[r=2]{c}{\textbf{11--50}\\(980)}
      & \textbf{MAE}  & 10.49 & \SetCell{font=\bfseries} 1.26 \\
      \SetHline[1]{2-4}{dashed}
      & \textbf{RMSE} & 12.52 & \SetCell{font=\bfseries} 2.52 \\

    \SetCell[r=2]{c}{\textbf{51--100}\\(377)}
      & \textbf{MAE}  & 13.84 & \SetCell{font=\bfseries} 2.68 \\
      \SetHline[1]{2-4}{dashed}
      & \textbf{RMSE} & 25.07 & \SetCell{font=\bfseries} 5.48 \\

    \SetCell[r=2]{c}{\textbf{101--1000}\\(127)}
      & \textbf{MAE}  & 23.55 & \SetCell{font=\bfseries} 7.03 \\
      \SetHline[1]{2-4}{dashed}
      & \textbf{RMSE} & 51.11 & \SetCell{font=\bfseries} 19.54 \\
  \end{tblr}
  \caption{Counting (testing) results on the full 10-category VisDrone-DET\cite{zhu2021detectionVisDrone} dataset. The best result on each row is marked in \textbf{bold}.}
  \label{table:visdrone10}

\end{table}

\subsection{Evaluation Metrics}
    As in DSACA \cite{xu2021dilated} and \textcite{michel2022class}, we compare methods on the macro-MAE and macro-RMSE metrics, based on testing data. Best model weights are chosen from macro-MAE score over the validation data:
    \begin{subequations}\label{eq:metrics}
        \allowdisplaybreaks[1]
        \begin{align}
            \text{macro-MAE}_i
            &= \frac{1}{C} \sum^C_{c=1} |\hat{n}_{i,c} - n_{i,c}|, \label{eq:metric_mMAE} \\[4pt]
            \text{macro-RMSE}_i
            &= \sqrt{ \frac{1}{C} \sum^c_{i=1} (\hat{n}_{i,c} - n_{i,c})^2}. \label{eq:metric_mRMSE}
        \end{align}
    \end{subequations}
    Assuming $C$ is the number of categories in a dataset, where the estimate count $\hat{n}_c$ is defined in \cref{eq:count_est}, and the ground-truth count $n_c$ in \cref{eq:count_gt}. Whilst DSACA \cite{xu2021dilated} report a ``Mean Squared Error (MSE)'' metric, their definition of MSE is equivalent to macro-RMSE, and we refer to their metrics as such. In our tables we refer to the mean macro-MAE and macro-RMSE metrics over the whole testing set, as in previous work. With a testing set of $N_\text{testing}$ samples, these can be defined as:
    \begin{subequations}\label{eq:metricsset}
        \begin{align}
            \text{MAE}
            &= \frac{1}{N_\text{testing}} \sum^{N_\text{testing}}_{i=1} \text{macro-MAE}_i, \label{eq:metricsset_mMAE} \\[4pt]
            \text{RMSE}
            &= \frac{1}{N_\text{testing}} \sum^{N_\text{testing}}_{i=1} \text{macro-RMSE}_i. \label{eq:metricsset_mRMSE}
        \end{align}
    \end{subequations}

\begin{table}
  \vspace{0.8em}
  \centering
  \begin{tblr}{width=\linewidth,
              colspec={Q[c,m] Q[c,m] |C C C C},
              hline{1,Z} = {2pt},
              hline{2,3} = {1pt},
              hlines}
    \SetCell[c=6]{c} VisDrone-DET (8-category) \\
    \SetCell[c=2]{c} {\textbf{Category}} &  & \SetCell{c, font=\bfseries} {{{DOPNet \cite{mingpeng2024DOPNet}}}} & \SetCell{c, font=\bfseries} {{{DSACA \cite{xu2021dilated}}}} & \SetCell{c, font=\bfseries} {{{YOLO 11x \cite{yolo11_ultralytics}}}} & \SetCell{c, font=\bfseries} {{{Ours}}} \\

    \SetCell[r=2]{c}{\textbf{All}}
      & \textbf{MAE}  & 3.48 & 3.43 & \uline{2.65} & \textbf{2.29} \\
      \SetHline[1]{2-6}{dashed}
      & \textbf{RMSE} & \uline{5.60} & \textbf{5.36} & 10.06 & 8.28 \\
      
    \SetCell[r=2]{c}{\textbf{People}}
      & \textbf{MAE}  & 8.63 & \textbf{5.04} & 10.00 & \uline{7.51} \\
      \SetHline[1]{2-6}{dashed}
      & \textbf{RMSE} & \uline{13.05} & \textbf{7.65} & 26.29 & 21.56 \\

    \SetCell[r=2]{c}{\textbf{Bicycle}}
      & \textbf{MAE}  & 2.34 & 2.35 & \uline{0.72} & \textbf{0.70} \\
      \SetHline[1]{2-6}{dashed}
      & \textbf{RMSE} & 4.34 & 4.33 & \uline{2.12} & \textbf{1.94} \\

    \SetCell[r=2]{c}{\textbf{Motor}}
      & \textbf{MAE}  & 4.48 & 8.90 & \textbf{2.18} & \uline{2.19} \\
      \SetHline[1]{2-6}{dashed}
      & \textbf{RMSE} & 6.55 & 12.23 & \uline{5.45} & \textbf{4.96} \\

    \SetCell[r=2]{c}{\textbf{Tricycle}}
      & \textbf{MAE}  & 2.54 & 2.88 & \textbf{0.43} & \uline{0.51} \\
      \SetHline[1]{2-6}{dashed}
      & \textbf{RMSE} & 4.21 & 4.61 & \uline{1.27} & \textbf{1.17} \\

    \SetCell[r=2]{c}{\textbf{Car}}
      & \textbf{MAE}  & 5.49 & 3.98 & \uline{3.77} & \textbf{3.07} \\
      \SetHline[1]{2-6}{dashed}
      & \textbf{RMSE} & 8.65 & \uline{6.02} & 7.53 & \textbf{5.37} \\

    \SetCell[r=2]{c}{\textbf{Van}}
      & \textbf{MAE}  & 2.57 & 2.54 & \uline{2.28} & \textbf{2.12} \\
      \SetHline[1]{2-6}{dashed}
      & \textbf{RMSE} & 4.57 & 4.51 & \uline{4.04} & \textbf{3.72} \\

    \SetCell[r=2]{c}{\textbf{Truck}}
      & \textbf{MAE}  & 1.36 & 1.32 & \textbf{0.98} & \uline{1.18} \\
      \SetHline[1]{2-6}{dashed}
      & \textbf{RMSE} & \uline{2.25} & 2.59 & \textbf{2.11} & 2.42 \\

    \SetCell[r=2]{c}{\textbf{Bus}}
      & \textbf{MAE}  & \uline{0.45} & \textbf{0.42} & 0.80 & 1.01 \\
      \SetHline[1]{2-6}{dashed}
      & \textbf{RMSE} & \uline{1.14} & \textbf{0.97} & 2.03 & 2.21 \\
    
  \end{tblr}
  \caption{Counting (testing) results on the merged 8-category VisDrone-DET \cite{zhu2021detectionVisDrone} dataset. A class-wise breakdown of performance is provided for each model, as DSACA \cite{xu2021dilated} does not provide range-wise breakdowns. The best results in each row are in \textbf{bold}, with the second-best results \uline{underlined}. The best metric is determined before rounding.\vspace{-1em}}
  \label{table:visdrone8}
\end{table}

    \section{Results and Discussion}
\label{sec:results}

We achieve a significant improvement in MAE across the 8 and 10 category VisDrone benchmarks (\cref{table:visdrone10,table:visdrone8}), and an even greater jump on the iSAID benchmark \cref{table:isaid}, attaining state-of-the-art RMSE metrics. To address the large gap in multi-class density estimation research, as DSACA \cite{xu2021dilated} was published in 2021, and \textcite{michel2022class} published their results in 2022, and to address other advances in object counting since then, we compare our results to the centroid prediction method DOPNet \cite{mingpeng2024DOPNet}, and against the largest size of YOLO11. \cref{table:ablations,table:isaid} demonstrate that YOLO11 performs poorly for extremely dense scenes (i.e. scenes with counts ranging 50--10000), and moderately in the ranges 0--10 and 10--50, evidencing the hypothesis that density estimation methods are better suited to dense counting applications than object-detection-based methods.

\subsection{Ablation Studies}

    In order to verify the significance of individual contributions, we ablate the CFM, the size of the backbone, and final-layer activation function in \cref{table:ablations}. \cref{table:ablations} demonstrates that the CFM improves overall model performance, especially at smaller count ranges. The effect of propagating the cross-entropy loss through the CFM is seen when we directly take the final layer of the backbone as the input to the CFM, in the single scale experiment. We see a change in the metrics compared to the full CFM, even when in all cases the output of the mask is discarded, and has no direct effect on the metrics at inference time. Whilst the \emph{0--1000} results of the single-scale experiment are very marginally improved over the base model, the model loses its scale independence, only predicting larger ranges of counts better than the base model with the multiscale CFM, and dropping in performance when predicting lower ranges of counts.

    The final-layer activation function is found to have a significant impact on the performance. Selecting leaky ReLU as a philosophical middle ground between the ReLU and softplus activations, we notice a sharp drop in low count ranges with the leaky variant. As the leaky ReLU is unbounded, we notice metrics being impacted from negative count predictions.

\begin{table}[h]
  \centering
  \begin{tblr}{width=\linewidth,
              colspec={Q[c,m] Q[c,m]|C C C},
              hline{1,Z} = {2pt},
              hline{2,3} = {1pt},
              hlines}
    \SetCell[c=5]{c} iSAID (4-category) \\
    \SetCell[c=2]{c} {\textbf{Count Range} \\ (samples in range)} & & {{{\textbf{YOLO 11x \cite{yolo11_ultralytics}}}}} & {{{\textbf{\citeauthor{michel2022class} \cite{yolo11_ultralytics}}}}} & {{{\textbf{Ours}}}} \\

    \SetCell[r=2]{c}{\textbf{0--10000}\\(4056)}
      & \textbf{MAE}  & 10.95 & 7.85 & \textbf{2.84} \\
      \SetHline[1]{2-5}{dashed}
      & \textbf{RMSE} & 68.20 & 31.64 & \textbf{29.12} \\
      
    \SetCell[r=2]{c}{\textbf{0--10}\\(2862)}
      & \textbf{MAE}  & 1.67 & 2.5 [\emph{sic}] & \textbf{1.04} \\
      \SetHline[1]{2-5}{dashed}
      & \textbf{RMSE} & 13.49 & \textbf{10.26} & 16.27  \\

    \SetCell[r=2]{c}{\textbf{11--50}\\(784)}
      & \textbf{MAE}  & 6.84 & 6.51 & \textbf{3.32} \\
      \SetHline[1]{2-5}{dashed}
      & \textbf{RMSE} & 19.61 & \textbf{12.18} & 28.07 \\

    \SetCell[r=2]{c}{\textbf{51--100}\\(201)}
      & \textbf{MAE}  & 20.15 & 17.86 & \textbf{6.54} \\
      \SetHline[1]{2-5}{dashed}
      & \textbf{RMSE} & 36.05 & 26.38 & \textbf{24.43} \\

    \SetCell[r=2]{c}{\textbf{101--10000}\\(209)}
      & \textbf{MAE}  & 144.53 & 50.31 & \textbf{22.11} \\
      \SetHline[1]{2-5}{dashed}
      & \textbf{RMSE} & 291.71 & \textbf{95.57} & 96.44 \\
  \end{tblr}
  \caption{Counting (testing) results on the reduced 4-category iSAID dataset \cite{waqas2019iSAID}, as described in \textcite{michel2022class}. The best result on each row is marked in \textbf{bold}. \vspace{-1em}}
  \label{table:isaid}
\end{table}

\subsection{Environmental Impact}
    This work required 2,439.00 GPU-hours (mix of NVIDIA V100, A100, RTX A6000, and RTX 3070), largely during methodological exploration, and mostly on the University of Lincoln (\emph{Novel}), and University of Aberdeen (\emph{Maxwell}) HPC clusters. These computations consumed 494.095 kWh of GPU energy, generating approximately 96.6 kg of CO\textsubscript{2}e emissions in England and Scotland \cite{webCO2Eqv2025}.

\begin{table}[h]
  \centering
  \begin{tblr}{width=\linewidth,
              colspec={Q[c,m] Q[c,m]|C C},
              hline{1,Z} = {2pt},
              hline{2,3} = {1pt},
              hlines}
    \SetCell[c=4]{c} \citeauthor{hicks2021deep} (8-category) \\
    \SetCell[c=2]{c} {\textbf{Count Range} \\ (samples in range)} & & {{{\textbf{DSACA \cite{xu2021dilated}}}}} & {{{\textbf{Ours}}}} \\

    \SetCell[r=2]{c}{\textbf{0--1000}\\(420)}
      & \textbf{MAE}  & 0.92 & \textbf{0.38}  \\
      \SetHline[1]{2-4}{dashed}
      & \textbf{RMSE} & 2.10 & \textbf{1.71}  \\
      
    \SetCell[r=2]{c}{\textbf{0--5}\\(252)}
      & \textbf{MAE}  & 0.61 & \textbf{0.17}  \\
      \SetHline[1]{2-4}{dashed}
      & \textbf{RMSE} & 0.91 & \textbf{0.82}  \\

    \SetCell[r=2]{c}{\textbf{6--10}\\(71)}
      & \textbf{MAE}  & 0.93 & \textbf{0.31}  \\
      \SetHline[1]{2-4}{dashed}
      & \textbf{RMSE} & 1.54 & \textbf{0.97}  \\

    \SetCell[r=2]{c}{\textbf{11--25}\\(71)}
      & \textbf{MAE}  & 1.39 & \textbf{0.69}  \\
      \SetHline[1]{2-4}{dashed}
      & \textbf{RMSE} & 2.78 & \textbf{2.01}  \\

    \SetCell[r=2]{c}{\textbf{26--1000}\\(26)}
      & \textbf{MAE}  & 2.64 & \textbf{1.76}  \\
      \SetHline[1]{2-4}{dashed}
      & \textbf{RMSE} & 5.96 & \textbf{5.22}  \\
  \end{tblr}
  \caption{Counting (testing) results on the reduced 8-category \textcite{hicks2021deep} dataset. The best result on each row is marked in \textbf{bold}.\vspace{-1em}}
  \label{table:hicks}
\end{table}
    \section{Conclusion}
\label{sec:conclusion}
In this work, we introduce a novel framework for multi-class density estimation, addressing the challenges of object counting in dense, occluded, and heterogeneous environments. By leveraging the Twins-SVT vision transformer backbone and a multiscale convolutional decoder, our method effectively captures both global context and fine-grained spatial details. The proposed Category Focus Module significantly reduces inter-class interference without the need for the application of secondary tasks at inference time, required in previous methods.
Extensive experiments on VisDrone, iSAID, and the biodiversity-focused Hicks dataset demonstrate that the method consistently outperforms existing multi-class counting methods and object detection baselines, achieving up to 64\% decrease in MAE. Furthermore, our method's applicability to ecological monitoring tasks highlights its potential for real-world applications beyond traditional urban scenarios.
We hope this work encourages further exploration into scalable, class-aware density estimation methods and fosters cross-domain innovation in automated counting systems.

    %---------------------------------------------------------------
    % Acknowledgment
    %---------------------------------------------------------------
    \section*{Acknowledgment}
    This work was supported by the UKRI AI Centre for Doctoral Training in Sustainable Understandable agri-food Systems Transformed by Artificial INtelligence (SUSTAIN) [grant reference: EP/Y03063X/1].

    %---------------------------------------------------------------
    % References
    %---------------------------------------------------------------
    \printbibliography

    %---------------------------------------------------------------
    % End of document
    %---------------------------------------------------------------
\end{document}